\documentclass[twoside,11pt]{article}

\usepackage{blindtext}

\usepackage{dmlr2e}
\usepackage{float}
\def\ourname{\emph{WarpSci}}
\newcommand*{\email}[1]{\texttt{#1}}

\usepackage{lastpage}
\dmlrheading{2}{2024}{1-\pageref{LastPage}}{2/2024;}{5/2024}{21-0000}{} % Insert the dates and author names. This is not relevant if it is yet being reviewed, i.e., \documentclass[twoside,11pt, preprint]{article}
\firstpageno{1}

\begin{document}

\title{Enabling High Data Throughput Reinforcement Learning on GPUs: A Domain Agnostic Framework for Data-Driven Scientific Research}

\author{\name {Tian Lan}$^{*}$, Huan Wang, Caiming Xiong and Silvio Savarese\\
       \addr Salesforce Research, USA \\
       \email tian.lan@salesforce.com}

\maketitle

\begin{abstract} We introduce \ourname{}, a domain agnostic framework designed to overcome crucial system bottlenecks encountered in the application of reinforcement learning to intricate environments with vast datasets featuring high-dimensional observation or action spaces. Notably, our framework eliminates the need for data transfer between the CPU and GPU, enabling the concurrent execution of thousands of simulations on a single or multiple GPUs. This high data throughput architecture proves particularly advantageous for data-driven scientific research, where intricate environment models are commonly essential.

\end{abstract}

\begin{keywords}
  throughput, reinforcement learning, gpu acceleration, data-driven science
\end{keywords}

\section{Introduction}

Reinforcement Learning (RL) stands out as a powerful algorithm for training AI agents, applicable in diverse domains such as strategy games\citep{OpenAI_dota, vinyals2019grandmaster}, robotics~\citep{gu2017deep, Ibarz_2021}, and large language models \citep{ouyang2022training}.  
Notably, there has been a recent surge in interest regarding the application of RL techniques in scientific research, encompassing diverse fields such as multi-agent\footnote{
An \emph{agent} is an actor in an environment. An \emph{environment} is an instance of a simulation and may include many agents with complex interactions. An agent is neither an environment nor a policy model.
} modeling in economics, climatology, and epidemiology \citep{zheng2022ai, trott2021building, zhang2022}; signal processing in astrophysics \citep{astronomy2022, yatawatta2023hint}; and investigating reaction paths in chemistry \citep{lan2021jacs, Yoon_2021}. However, numerous engineering and scientific challenges persist in the adoption of RL in scientific investigations. The performance of RL implementations can decelerate significantly when simulations become data-intensive, particularly in scenarios involving numerous agents or high-dimensional state or action spaces, resulting in experiments that span weeks. The comparatively low data throughput of RL further contributes to the emergence of non-stationary and strongly correlated data sequences, while the finite-horizon roll-out in RL introduces bias over the value function estimation \citep{mnih2016asynchronous, zhang2020global, lan2021jacs}. Regrettably, such complexity and challenges are commonplace in data-driven scientific modeling. For instance, in economic simulations, the construction of a realistic environment necessitates hundreds of agents and numerous actions \citep{zhang2022}. Similarly, the study of catalytic reaction pathways involves navigating a chemical potential energy landscape that can easily exceed twenty dimensions with extreme noise \citep{lan2021jacs}. While distributed systems are employed to scale RL performance, the associated costs of worker communication and data transfer can be very high \citep{espeholt2018impala, espeholt2020seed, hoffman2020acme, pretorius2021mava}, as detailed in Appendix \ref{app:scalableRL}.

\section{Contribution}
The primary objective of this \emph{Extended Abstract} is to bring attention to the challenge of RL in scientific research arising from the data throughput, and introduce our comprehensive solution, \ourname{}. \ourname{} is a computational framework specifically designed to achieve massively high-throughput and domain-agnostic RL simulation in the context of data-driven scientific research. The framework builds upon the foundation of WarpDrive \citep{JMLR:v23:22-0185} which is accessible at \url{https://github.com/salesforce/warp-drive}. 

\ourname{} performs the entire RL workflow on a single or multiple GPUs, utilizing a unified and in-place data store within GPUs for simulation roll-outs and training. This minimizes the data transfer between CPU and GPU or within GPU, reducing simulation and training time significantly. The framework also leverages GPU parallelization to concurrently run thousands of RL simulations, operating independently in the dedicated GPU blocks and concurrently producing exceptionally large batches of experience. 
\ourname{} offers simple Python classes located on the CPU to streamline all relevant CPU-GPU communication and interactions essential for RL, and offer simple toolings for constructing custom RL environments connected to the CUDA back-end.

This high throughput yet cost-effective architecture proves particularly advantageous for data-driven scientific research, where enormous data consumption, complex agent interactions, and diverse environments are usually indispensable. More details of the design choice and the computational architecture are provided in Appendix \ref{app:WarpSci-detail}.

\section{Examples}
We present three examples: \emph{gym} classic control \citep{gym-env} for benchmarking, a multi-agent economic simulation \citep{trott2021building}, and generalizable catalytic reaction paths modeling \citep{lan2021jacs, lan2024}. All experiments ran on a single Nvidia A100 GPU on the Google Cloud Platform. Due to space constraints, we provide a brief summary in this section, with more information in Appendix \ref{app:examples}.

\textbf{Throughput}: \ourname{} achieves significantly higher (at least $10-100\times$) throughput than the distributed systems at low cost (a single A100 GPU). For example, 8.6M environment steps/second for 10K concurrent cartpole environments, 0.12M for 1K concurrent economic simulations and 0.95M for catalytic reaction modeling with 2K concurrent environments \footnote{In certain experiments, we employed a reduced level of concurrency to optimize the trainer's capacity and memory space.}.  
Scaling almost linearly to thousands of environments or agents, \ourname{} demonstrates near-perfect parallelism. It can also train across multiple GPUs for further throughput scaling.
\textbf{Convergence}: Our study indicates that training with an increased data throughput generated by concurrent environments achieves faster and more stable global convergence. \textbf{Environments Agnostic}: \ourname{} offers tools to develop custom environments for diverse scientific research topics, and supports actor-critic algorithms for both discrete and continuous actions.

% Acknowledgements and Disclosure of Funding should go at the end, before appendices and references

\vskip 0.2in
\bibliography{main}

\newpage
\appendix
\section{Scalable Reinforcement Learning} 
\label{app:scalableRL}

Common scalable RL systems often employ a combination of distributed roll-out and trainer workers. Roll-out workers execute the environment to produce roll-outs, utilizing actions sampled from policy models on either roll-out workers or trainer workers. Typically, roll-out workers operate on CPU machines, occasionally utilizing GPU machines for richer environments.\citep{pretorius2021mava, hoffman2020acme, espeholt2018impala}.
Trainer workers gather roll-out data asynchronously from roll-out workers and iteratively optimize policies on either CPU or GPU machines.
While such a distributed design is scalable, worker communication and data transfer cost is expensive and individual machine utilization can be poor.
To improve performance, GPU and TPU-based RL frameworks exist \citep{tang2022evojax, hessel2021podracer}, but have focused on single-agent and domain-specific environments, e.g., for Atari \citep{dalton2020accelerating}, or learning robotic control in 3-D rigid-body simulations \citep{freeman2021brax, makoviychuk2021isaac}. 
Consequently, building efficient RL pipelines for simulations with intricate agent interactions, substantial data consumption, and diverse environments, as usually seen in scientific research, remains a challenging endeavor.

\section{Details of Architecture} 
\label{app:WarpSci-detail}

\begin{figure*}[h]
\centering
    \includegraphics[width=0.9\linewidth]{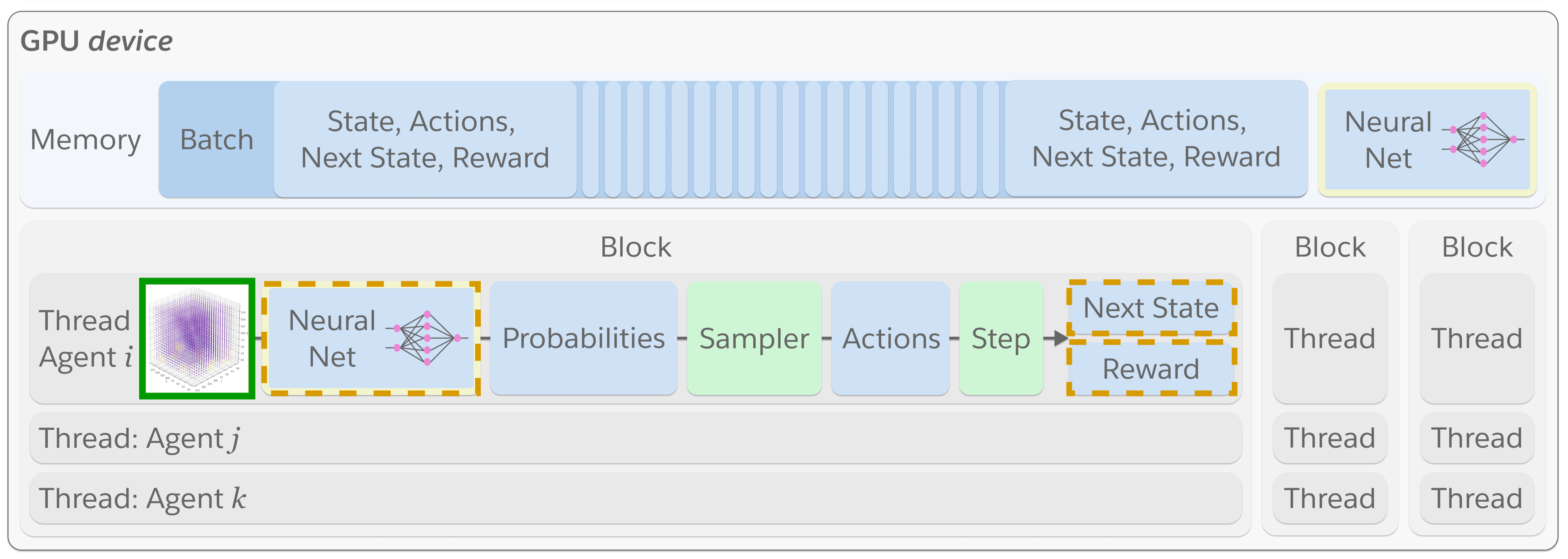}
    \caption{
    A flow chart depicting \emph{WarpSci}. Computations within this framework are organized into GPU blocks, each comprising multiple threads to facilitate concurrent environment roll-outs. Each thread is responsible for operating an agent that samples actions and computes rewards. These blocks have access to the global GPU memory, which houses the RL environment (depicted as a 3-D grid in a green-bordered box) with local variations, and deep policy models. Additionally, they store in-place roll-out data for training purposes. The dashed brown boxes represent references (not copies) of the policy model objects and data placeholders managed by blocks and hosted in the global memory. Users have the flexibility to compose and upload their custom environment setups to finalize the environment construction. 
    }
    \label{fig:warpsci-code-structure}
\end{figure*}

As shown in Fig. \ref{fig:warpsci-code-structure}, \ourname{} executes the entire RL workflow seamlessly on a single GPU or multiple GPUs, utilizing a unified data storage hosted within the GPU for simulation roll-outs, action inference, reset and training. This approach minimizes CPU-GPU data communication and eliminates the need for additional data transfer within the GPU, resulting in a substantial reduction in both simulation and training times. Furthermore, our framework achieves parallelization at low cost by concurrently running thousands of single-agent or multi-agent simulations, capitalizing on the inherent parallel processing capabilities of GPUs.
Each environment instance operates independently within a dedicated GPU block. Within each block, individual agents run on unique GPU threads, enabling interactions across threads. Each instance maintains a reference (not a copy) to the environment with local variations or random configurations, significantly reducing the storage overhead associated with the environment setup.

\ourname{} offers simple Python classes located on the CPU to streamline all relevant CPU-GPU communication and interactions essential for RL. These classes connect to the CUDA back-end and offer simple APIs for constructing high-level Python applications. Users only need supply the \emph{step} function to finalize the custom environment definition. As a default environment composer, we employ Numba, a user friendly, just-in-time compiler for Python. Finally, our framework automatically loads and integrates the environment \emph{step} into the environment-agnostic CUDA backend for the RL simulation.

\section{Example Details}
\label{app:examples}

All experiments ran on a single Nvidia A100 GPU, \emph{a2-highgpu-1g}, on the Google Cloud Platform.

\paragraph{Classic Control.}

In the field of RL, classic control environments usually serve as fundamental benchmarks to evaluate the performance of various RL algorithms and systems. These environments typically involve simple physics-based systems, yet their challenges lie in achieving stable and optimal control. Iconic examples, such as CartPole and Acrobot in \emph{gym} environment\citep{gym-env}, offer controlled scenarios with well-defined dynamics, making them ideal for benchmarking the throughput scalability and the learning capability of \ourname{}.

\begin{figure}[H]
\centering
    \includegraphics[width=0.55\linewidth]{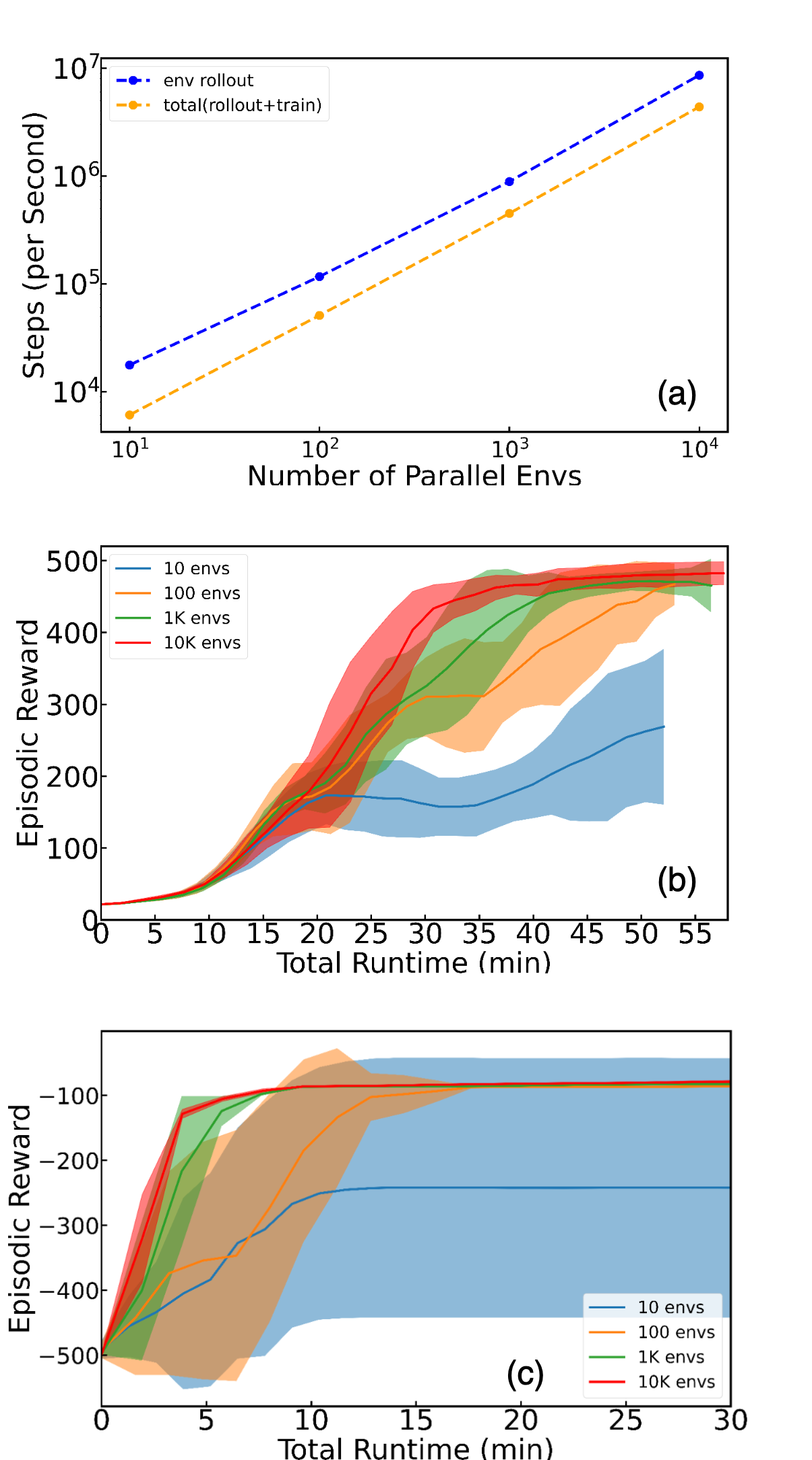}
    \caption{Scalability, convergence and learning speed for \ourname{} applied to \emph{gym} classic control environments. (a) Roll-out and training throughput in Cartpole-v1 and Acrobot-v1 versus the number of parallel environments (log-log scale) to 10K concurrent environments with random local initialization: the throughput scales linearly. 
    The average episodic reward (the accumulated total reward collected from the start to the terminal state)
    versus the training time (wall-clock minutes) for (b) Cartpole-v1 and (c) Acrobot-v1 running at various concurrency levels.
    The model was trained on a single Nvidia A100 GPU. For robustness, the depicted results are averaging over eight independent runs from scratch with different initialization seeds and the same hyperparameters. The shadow regions represent the error bar of eight independent runs. }
    \label{fig:classic-control}
\end{figure}

Fig. \ref{fig:classic-control}(a) shows that \ourname{}’s performance in classic control environments scales linearly to 10K of
environments, yielding perfect parallelism. For example, \ourname{} runs at 8.6 million environment steps per second with
10K Cartpole-v1 or Acrobot-v1 environments. Fig. \ref{fig:classic-control}(b) and (c) displays the convergence speed of \ourname{} as a function of the number of environment replicas running in parallel. The data reveal that, under consistent fixed hyperparameters, the simulations operating with an increased number of concurrent environments attain global convergence faster and more stably. Particularly, simulations with 10K Cartpole and Acrobot environment replicas reach the global optimum within 30 and 5 minutes respectively, while 10 environment replicas can barely exhibit satisfactory convergence in such a short period. 

\paragraph{Multi-Agent Economics.}

\begin{figure*}[t!]
\centering
    \includegraphics[width=0.75\linewidth]{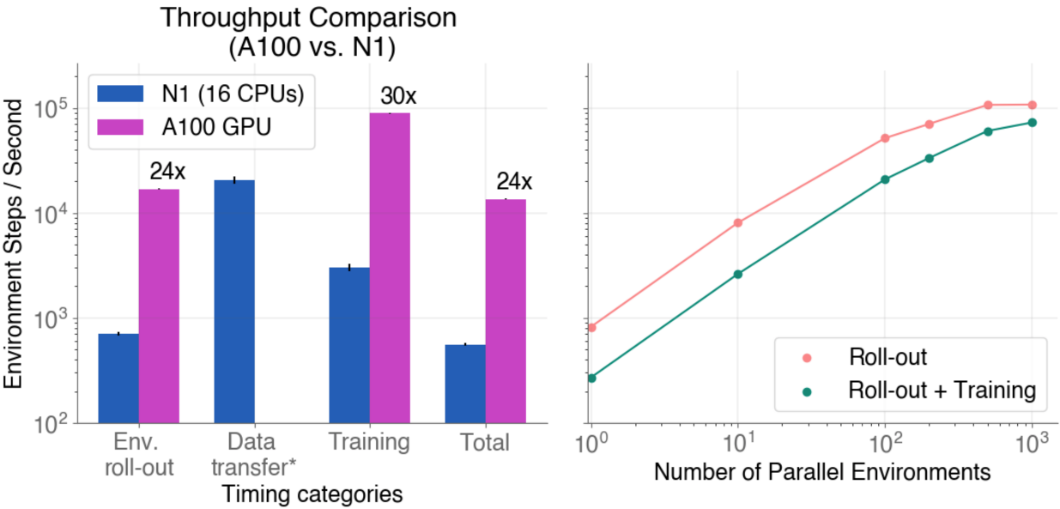}
    \caption{\ourname{} performance in the COVID-19 economic simulation in log scale. 
    Left: Note that there is no data transfer with \ourname{}. With 60 parallel environments, \ourname{} achieves 24 times higher throughput over CPU-based distributed training architectures (“total”). Moreover, both the roll-out and training phase are an order of magnitude faster than on the distributed N1 node. 
    Right: Environment steps per second and end-to-end training speed scale
    almost linearly with the number of environments. (Credit: \cite{lan2021warpdrive})
    }
    \label{fig:economics}
\end{figure*}
We demonstrate the scalability of \ourname{} to more intricate environments through its evaluation in a COVID-19 simulation. This simulation, grounded in real-world data, models the interplay between health and economic dynamics during the COVID-19 pandemic. Notably, the simulation step is significantly more complex compared to the \emph{gym} classic control problems, consuming a larger fraction of each iteration's runtime.

The simulation involves 52 agents, with 51 representing governors for each U.S. state and Washington D.C., and an additional agent for the federal government of the USA. This constitutes a complex two-level multi-agent environment, where state agents determine the stringency level of the pandemic response, and the federal government provides subsidies to eligible individuals. The actions of each agent influence health and economic outcomes, such as deaths, unemployment, and GDP. Moreover, the federal government's actions can alter the health-economic trade-off and optimization objective for the U.S. states, rendering it a complex and dynamic two-level RL problem. Interested readers seeking additional scientific background and technical details are encouraged to refer to \cite{trott2021building, zheng2022ai}.

For this study, \ourname{} achieves 24 times higher throughput with 60 environment replicas, compared to a 16 CPU
node, \emph{n1-standard-16}, on the Google Cloud Platform. Across different timing categories as shown in Fig. \ref{fig:economics},
the performance gains comprise a 24 times speed-up during the
environment roll-out, a zero data transfer time, and a 30 times
speed-up for training the policy models. 
Moreover, \ourname{} can scale almost linearly to 1K parallel COVID-19
environments, resulting in even higher throughput gains.

\paragraph{Catalytic Reactions.}

\begin{figure*}[t!]
\centering
    \includegraphics[width=1.0\linewidth]{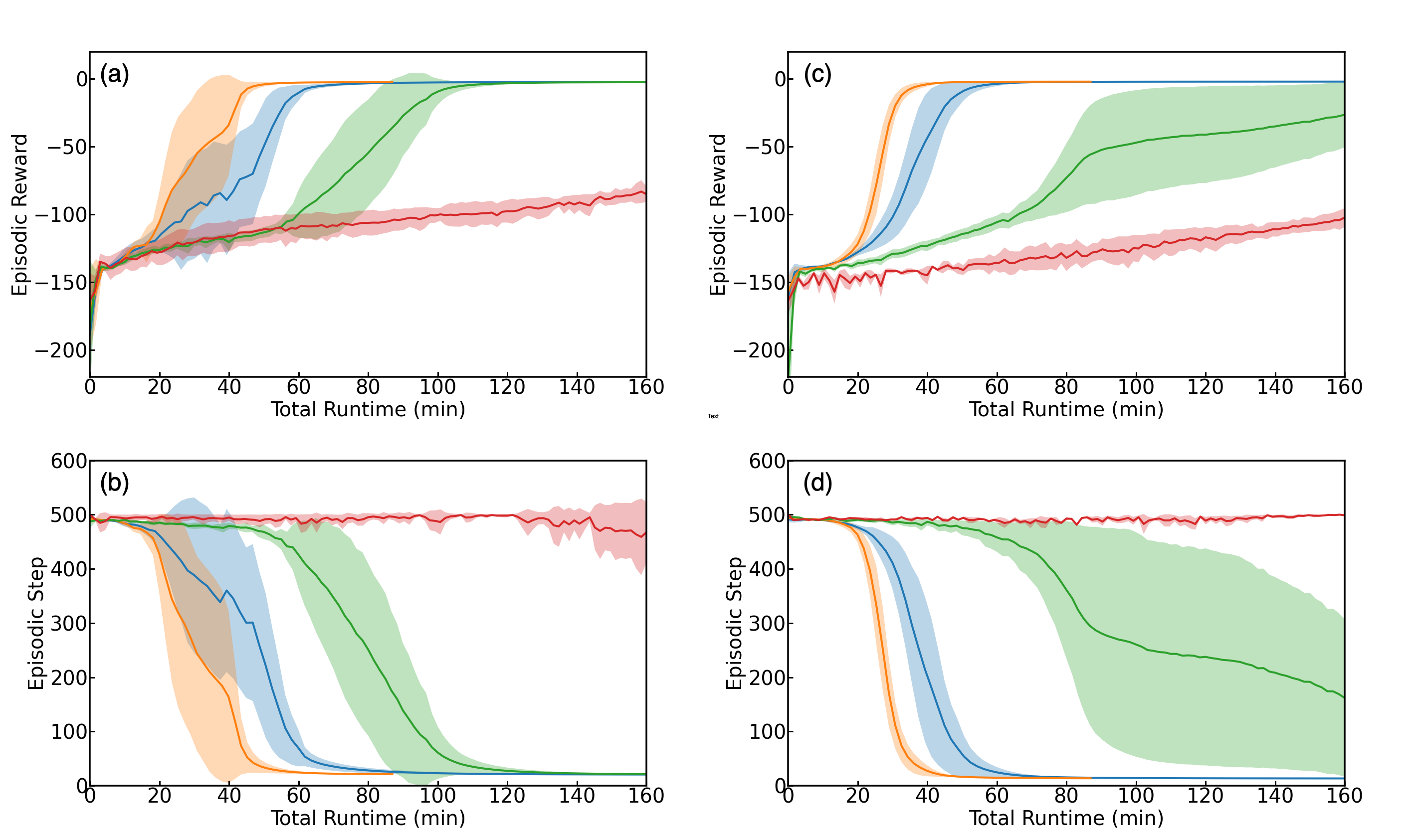}
    \caption{Convergence and learning speed, quantified by total runtime in wall-clock minutes, were assessed for Langmuir-Hinshelwood (a, b) and Eley-Rideal (c, d) hydrogenation reactions of NH$_2$ to NH$_3$. Varied numbers of concurrent environment instances were employed: 4 in red, 20 in green, 100 in blue, and 500 in yellow. The episodic reward denotes the mean accumulated reward that H atom actors gather from the initial to the terminal state of (a) Langmuir-Hinshelwood and (c) Eley-Rideal. Episodic step indicates the average total steps to reach the terminal state for (b) Langmuir-Hinshelwood and (d) Eley-Rideal. Training utilized a single Nvidia A100 GPU. For robustness, the displayed results are averages over five independent runs from scratch with different initialization seeds and identical hyperparameters. Shadow regions depict the error bars of the five independent runs. (Credit: \cite{lan2024})}
    \label{fig:catalytic}
\end{figure*}
Comprehending catalytic reaction pathways is essential for advancing our understanding of chemical processes, refining conditions, and designing robust catalysts. These pathways offer insights into reaction mechanisms, facilitating the creation of more selective catalysts \citep{chem-rev1, chem-rev2}. However, exploring these pathways presents significant challenges, including the complexity of multi-step reactions, short-lived intermediates, and experimental intricacies \citep{chem-rev3, jacsau, lan2021jacs}. RL shows promise in overcoming these challenges by providing an automated approach to navigating reaction networks. However, RL encounters scientific and engineering obstacles, primarily limited by simulation throughput. Consequently, current RL research in chemical reactions often concentrates on specific reactions, relying on model simplifications with state vector encodings or heuristic rules. This approach limits generalizability and requires substantial empirical design. Exploration is also confined to predefined sets of reaction networks, hindering the discovery of unknown mechanisms \citep{Yoon_2021, lan2021jacs, nature-cat}. Therefore, the pursuit of a more versatile RL solution to explore undiscovered reaction mechanisms remains a significant challenge in the field.

In this study, we present a reaction-agnostic methodology facilitated by \ourname{}. The RL environment is constructed solely based on the potential energy landscape derived from first principles. This approach intrinsically defines the chemical reaction environment as a function of atomic positions, eliminating the necessity for laborious empirical or semi-empirical design of reaction-specific representations in RL environments. The outstanding generalizability and training speed are supported by the remarkable high-throughput capacity enabled by our architecture. 

We forecast the reaction pathway for the crucial hydrogenation step in the Haber-Bosch (H-B) process on the Fe(111) surface. The H-B process holds a pivotal role in Earth’s nitrogen cycle and represents over 2 percent of global energy consumption, yielding 160 million tons of ammonia annually. Despite a century of concentrated research to improve the H-B process, progress has been slow \citep{Chen2018}. Our framework has the potential to significantly contribute to process optimization, potentially reducing production costs and CO$_2$ emissions while enabling the establishment of smaller and more widespread plants.

Figure \ref{fig:catalytic} displays the convergence speed as \ourname{} processes the Langmuir-Hinshelwood reaction as a function of the number of environment replicas, running in parallel. The data reveal that, under consistent fixed hyperparameters, the simulations operating with an increased number of concurrent environments attain global convergence faster and more stably. The generalizable RL environment with the same hyperparameters is directly applicable to the study of Eley-Rideal reaction mechanism. The results highlight the critical role of massively high data throughput in RL for effectively exploring a broad range of reaction mechanisms through a generalizable RL environment representation built solely upon atomic positions.

Our findings reveal that the Langmuir-Hinshelwood mechanism shares the same transition state as the Eley-Rideal mechanism for H migration to NH2, forming ammonia. Furthermore, the reaction path identified by our model exhibits a lower energy barrier compared to that through nudged elastic band calculation. In this \emph{Extended Abstract}, we focus on presenting the generalizability, training speed and convergence stability facilitated by the high throughput of \ourname{}.
Interested readers seeking additional scientific background and technical details of this study are encouraged to refer to \cite{lan2021jacs, lan2024, nature-cat}.

\end{document}